\documentclass[journal]{IEEEtran}

\usepackage{amsmath,amssymb,amsfonts}   
\usepackage{amsthm}                      

\usepackage{booktabs}        
\usepackage{array}           
\usepackage{multirow}        
\usepackage{threeparttable}  
\usepackage{makecell}        
\usepackage{siunitx}         

\usepackage{graphicx}
\usepackage[caption=false,font=normalsize]{subfig}  
\usepackage{stfloats}        

\usepackage{algorithm}
\usepackage{algorithmic}

\usepackage{textcomp}
\usepackage{url}             
\usepackage{verbatim}        
\usepackage{cite}            

\usepackage{xspace}          

\newtheorem{theorem}{Theorem}

\begin{document}

\title{Communication-Efficient Multi-Modal Edge Inference via Uncertainty-Aware Distributed Learning}

\author{\IEEEauthorblockN{Hang Zhao, Hongru Li, Dongfang Xu, Shenghui Song, and Khaled B. Letaief,~\textit{Fellow,~IEEE}}
\\\IEEEauthorblockA{\textit{Dept. of Electronic and Computer Engineering,
The Hong Kong University of Science and Technology, Hong Kong}\\
E-mail: hzhaobi@connect.ust.hk, hlidm@connect.ust.hk, \\
eedxu@ust.hk, eeshsong@ust.hk, eekhaled@ust.hk}\\ \vspace*{-12mm}\thanks{This work was presented in part at the IEEE International Mediterranean Conference on Communications and Networking (MeditCom), July 2025~\cite{zhao2025multi}. }
}

\maketitle
\begin{abstract}
Semantic communication is emerging as a key enabler for distributed edge intelligence due to its capability to convey task-relevant meaning. However, achieving communication-efficient training and robust inference over wireless links remains challenging. This challenge is further exacerbated for multi-modal edge inference (MMEI) by two factors: 1) prohibitive communication overhead for distributed learning over bandwidth-limited wireless links, due to the \emph{multi-modal} nature of the system; and 2) limited robustness under varying channels and noisy multi-modal inputs. In this paper, we propose a three-stage communication-aware distributed learning framework to improve training and inference efficiency while maintaining robustness over wireless channels. In Stage~I, devices perform local multi-modal self-supervised learning to obtain shared and modality-specific encoders without device--server exchange, thereby reducing the communication cost. In Stage~II, distributed fine-tuning with centralized evidential fusion calibrates per-modality uncertainty and reliably aggregates features distorted by noise or channel fading. In Stage~III, an uncertainty-guided feedback mechanism selectively requests additional features for uncertain samples, optimizing the communication--accuracy tradeoff in the distributed setting. Experiments on RGB--depth indoor scene classification show that the proposed framework attains higher accuracy with far fewer training communication rounds and remains robust to modality degradation or channel variation, outperforming existing self-supervised and fully supervised baselines.
\end{abstract}


\section{Introduction}

\IEEEPARstart{T}{he} advent of sixth-generation (6G) wireless networks is set to revolutionize the delivery of intelligent services, such as immersive extended reality (XR), digital twins, autonomous driving, and large-scale Internet of Things (IoT) ecosystems~\cite{8808168, zhu2020toward}. Many of these applications rely on distributed edge intelligence, where multiple sensing devices and edge/cloud servers collaboratively perform learning and inference over bandwidth-limited and interference-prone wireless links. These services differ significantly from traditional data services, which prioritize reliable bit-level transmission and reconstruction at the receiver~\cite{shannon1948mathematical}. Instead, they focus on transmitting relevant information for downstream tasks such as object detection and tracking. This shift from bit-level accuracy to task-level efficiency has led to the emergence of \emph{task-oriented} or \emph{semantic} communication~\cite{lu2022rethinking,  gunduz2022beyond, luo2022semantic, yang2022semantic} as a promising paradigm for distributed learning and inference over communication-constrained networks. By leveraging deep neural networks (DNNs) to create end-to-end trainable models that map high-dimensional sensor data into compact, task-relevant representations, semantic communication can significantly reduce the communication burden over bandwidth-constrained channels while preserving or even enhancing task performance~\cite{weng2021semantic, shao2021learning}.

Early works on semantic communication mainly focused on single-modality inference. The research on deep joint source–channel coding (JSCC)~\cite{bourtsoulatze2019deep} showed that directly mapping data to channel symbols with neural networks can achieve robust reconstruction over a wide range of channel conditions, outperforming separate source and channel coding. Subsequent studies adopted a task-oriented perspective, where the transmitter extracts and sends task-relevant features for downstream objectives such as image classification, object detection, semantic segmentation, speech recognition, and natural language processing~\cite{xie2021deep, shao2021learning, jiang2022deep, weng2023deep, kurka2020deepjscc, kurka2021bandwidth}. This line of work significantly improves inference-time communication efficiency and end-task performance. 
However, most systems are trained in a centralized manner and only focus on reducing the transmission cost for inference. They largely overlook the communication overhead of distributed learning where the transmitter and receiver are physically separated and connected via bandwidth-limited wireless links.
This issue is further amplified in distributed edge intelligence architectures, where many devices must continually adapt their encoders and decision modules to non-stationary data and varying channels.

To support richer sensing and more complex tasks, recent works have extended semantic communication to multi-modal settings~\cite{zhang2024unified, du2025task, jiang2024large, xie2022task, li2022cross, luo2022multimodal, xie2023semantic}. These methods exploit cross-modal redundancy and complementarity to enhance task performance and robustness, e.g., via unified multi-task multi-modal systems with dynamic feature selection and shared codebooks~\cite{zhang2024unified}, large-multi-modal-model-based frameworks for vehicular and general wireless networks~\cite{du2025task, jiang2024large}, and feature-level fusion or cooperative schemes across devices and channels~\cite{xie2022task, li2022cross, luo2022multimodal, tian2025synchronous, zhou2025multi, shao2022task}. In distributed edge intelligence applications, multiple spatially separated devices collect heterogeneous sensor data (e.g., RGB, depth, LiDAR, audio), locally encode them into semantic features, and send these features to an edge server for fusion and decision making. Typical applications include autonomous driving and smart healthcare scenarios~\cite{du2025task}. By transmitting distilled task-relevant semantics, these systems achieve high inference accuracy under stringent bandwidth constraints.

Despite these advances, most multi-modal semantic communication designs focus on the inference stage~\cite{zhou2025multi, shao2022task}. Specifically, they typically assume that encoders and fusion modules are trained centrally or offline, and overlook the communication cost of distributed learning over the network. Moreover, they pay limited attention to the issues caused by varying wireless channels and noisy modalities, for both training and inference. These limitations call for communication-aware frameworks that jointly reduce training- and inference-phase overhead while maintaining reliable multi-modal decision making. However, realizing such communication-aware frameworks in realistic edge systems gives rise to several key challenges in terms of communication efficiency, robustness, and system scalability, as shown below.

\textbf{Challenge 1 – Communication-efficient distributed training and model updates:}
In practical edge intelligence systems, device-side encoders must be updated to accommodate non-stationary data, evolving tasks, and changing channels. This entails iterative exchanges of model parameters, gradients, or intermediate representations between devices and the edge. Prior studies in the single-modal case show that such training-time synchronization can dominate the overall communication budget~\cite{li2025remote}. In the multi-modal distributed setting, each device typically hosts a modality-specific encoder. Jointly training these encoders with the edge-side fusion module requires repeated \emph{cross-device} coordination under per-device communication budgets and shared wireless uplink resources. Consequently, the objective is not merely to compress generic training traffic, but also to design distributed multi-modal training schemes that explicitly control the number of communication rounds and per-round payload sizes while enabling effective joint representation learning and high downstream accuracy.

\textbf{Challenge 2 – Robustness under varying channels and noisy multi-modal inputs:}
During deployment, fluctuating wireless channels and modality-specific sensing distortions can lead to corrupted, delayed, or missing features from different devices. Although robustness has been studied in single-modality semantic communication~\cite{peng2022robust, hu2022robust, nan2023physical, hu2023robust, peng2024robust}, multi-modal robustness under data noise and varying wireless channels remains largely unexplored. In practice, different modalities suffer from distinct distortions~\cite{garcia2018modality, lan2019learning, ma2021trustworthy} (e.g., poor illumination for RGB, occlusions for depth/LiDAR, and acoustic interference for audio), while device-specific channel states can cause some modalities to be frequently dropped, strongly compressed, or severely delayed. These factors complicate robust multi-modal fusion at the edge/server and can cause dramatic performance drops when the fusion module faces \emph{ imbalanced and partially missing} features. Achieving reliable inference in this regime requires the joint design of robust semantic encoders at distributed devices and noise-resilient multi-modal fusion at the edge/server.

\par
To the best of the authors' knowledge, no prior work has jointly addressed the above two challenges in a unified manner for multi-modal semantic communication, especially in distributed edge inference systems where multiple devices transmit semantic features to an edge server over wireless links. In this setting, representation learning, model updating, and decision making must be coordinated across spatially separated devices under varying communication conditions. Simply ``pushing'' a centralized design to distributed devices ignores per-device budgets and continual model updates. Instead, practical edge systems require a unified framework that learns communication-efficient multi-modal encoders, performs robustness-aware fusion that accounts for sensing quality and channel states, and adaptively controls inference-time communication via feedback or retransmissions.

In this paper, we propose a unified three-stage training and inference framework for distributed multi-modal edge inference that jointly optimizes communication efficiency across the model lifecycle and improves robustness under realistic wireless conditions. Unlike most prior multi-modal semantic communication designs that primarily optimize inference with centrally trained models and offer limited principled treatment of channel/noise effects, the proposed framework is fundamentally new in three aspects: (i) it establishes information-theoretic guarantees for multi-modal self-supervised learning under channel constraints, characterizing how task-relevant information is preserved despite augmentation and channel-induced distortions; (ii) it introduces a decoupled cross-modal and intra-modal pre-training loss that is fully executable locally at devices with zero device–server exchange, yielding shared and modality-specific encoders that reduce labeled data requirements and substantially cut training communication rounds; and (iii) it develops a distributed quantile-based uncertainty-guided retransmission policy built on evidential uncertainty, enabling an explicit and controllable communication–accuracy tradeoff at inference time by requesting additional features only for highly uncertain samples or adverse channel states.
Experiments on RGB–depth indoor scene classification under varying wireless channels and noisy inputs validate these advantages. The proposed framework attains higher accuracy with far fewer device–server communication rounds and remains robust even when one modality is severely degraded, outperforming existing self-supervised and fully supervised semantic communication baselines.

\par
The main contributions are summarized as follows:
\begin{itemize}
\item \textbf{Three-stage distributed framework.} We cast multi-modal semantic communication as distributed learning and inference under joint training/inference communication constraints, and propose a lifecycle-wide three-stage design. This includes local self-supervised pre-training on devices, server-side supervised fine-tuning with evidential fusion, and uncertainty-guided adaptive communication at inference, improving both efficiency and robustness under dynamic channels and degraded modalities.

\item \textbf{Local multi-modal self-supervised pre-training.} In Stage~I, we develop a fully local objective that disentangles shared and modality-specific representations and enhances intra-/cross-modal structure, yielding a zero-communication initialization. Theory and experiments show improved label-relevant information preservation and channel robustness, reducing labeled-data demand and downstream device--server communication rounds.

\item \textbf{Evidential fusion and uncertainty-guided retransmission.} 
In Stage~II, we equip each modality with an evidential head that converts the \emph{received} semantic features into Dirichlet evidence and performs reliability-aware fusion at the server, yielding better-calibrated epistemic uncertainty for both per-modality and fused predictions \emph{without increasing the transmitted feature payload}. In Stage~III, the server leverages these uncertainty estimates to trigger feedback-driven retransmissions only when needed: for modality--sample pairs whose uncertainty exceeds a quantile-calibrated threshold, the server requests additional feature transmissions up to $N_{\max}$. This policy requests additional features only for high-uncertainty samples, achieving a controllable communication--accuracy tradeoff.

\item \textbf{Wireless edge validation.} 
On distributed RGB--depth indoor scene classification under varying signal-to-noise ratios, fading/noise impairments, and corrupted multi-modal inputs, the proposed self-supervised pre-training reaches target accuracy with about \,$10\times$ fewer training communication rounds than contrastive pre-training and more than \,$20\times$ fewer than training from scratch, while also improving final accuracy. 
During inference, evidential fusion combined with uncertainty-guided retransmission consistently outperforms a JSCC baseline and remains robust when one modality is severely degraded, achieving higher accuracy with only a small fraction of additional transmissions on average.
\end{itemize}

The rest of this paper is organized as follows. Section~II presents the system model and problem formulation. Section~III describes the three-stage framework. Section~IV reports experiments and ablations. Finally, the conclusions are given in Section V.

Notations: Scalars are denoted by lowercase letters (e.g., $x$), and \emph{random scalars} by uppercase letters (e.g., $X$).
Vectors and matrices use boldface; random vectors/matrices are in bold uppercase (e.g., $\boldsymbol{X}, \boldsymbol{A}$) and their realizations are in bold lowercase (e.g., $\boldsymbol{x}, \boldsymbol{a}$).
Sets/spaces use calligraphic letters (e.g., $\mathcal{X}$), and probability measures/distributions use blackboard bold (e.g., $\mathbb{P}$).
The real and complex matrix spaces are $\mathbb{R}^{N\times M}$ and $\mathbb{C}^{N\times M}$.
The Euclidean norm is $\lVert \cdot \rVert_2$ (squared norm $\lVert \cdot \rVert_2^{2}$), and the inner product is $\langle \cdot, \cdot \rangle$.
Expectation and variance are $\mathbb{E}[\cdot]$ and $\mathrm{Var}[\cdot]$; the Kullback–Leibler divergence is $\mathrm{KL}(\cdot\Vert\cdot)$.
The mutual information between $X$ and $Y$ is $I(X;Y)$, and the conditional mutual information is $I(X;Y\mid Z)$.
When we write $I(X^{1};X^{2};Y)$ we refer to the \emph{interaction information} $I(X^{1};X^{2}) - I(X^{1};X^{2}\mid Y)$.
For multi-modal variables we index the modality by superscripts, e.g., $\boldsymbol{X}^{1}$ (\textsc{rgb}) and $\boldsymbol{X}^{2}$ (\text{depth});
the concatenation $[\boldsymbol{X}^{1};\boldsymbol{X}^{2}]$ is denoted $\mathrm{concat}(\boldsymbol{X}^{1},\boldsymbol{X}^{2})$ or, when clear from context, $\boldsymbol{X}^{1:2}$.

\section{System Model and Problem Formulation}
\subsection{System Model}
\begin{figure}[t]
    \centering
    \includegraphics[width=0.5\textwidth]{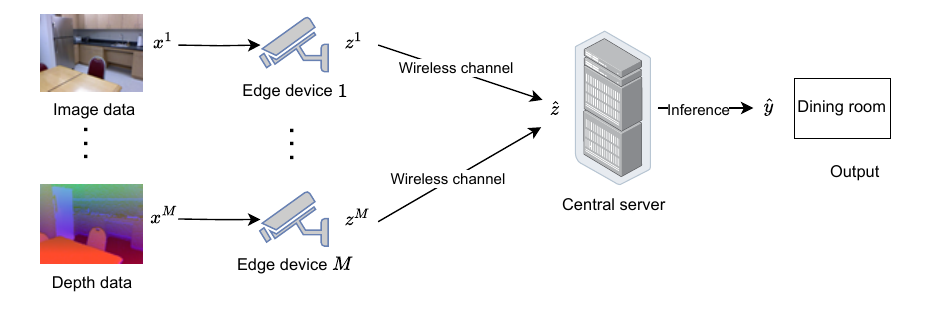}
    \caption{Multi-modal semantic communication for edge--server collaborative inference.}
    \label{framework}
\end{figure}

As shown in Fig.~\ref{framework}, we consider a distributed multi-modal semantic communication framework for collaborative inference in classification tasks, over bandwidth-limited wireless links.
The system consists of $M$ edge devices and a central server. Each device is equipped with a sensor and a modality-specific semantic encoder. At device $m\in\mathcal{M}\!=\!\{1,\ldots,M\}$, the sensor acquires raw data $\boldsymbol{x}^m \!\in\! \mathbb{R}^{D_m}$ and the encoder maps $\boldsymbol{x}^m$ to a semantic feature vector $\boldsymbol{z}^m$.
For simplicity, we assume that all modality-wise semantic features have the same dimension, i.e., $\boldsymbol{z}^m \in \mathbb{R}^{K}$ for all $m\in\mathcal{M}$.
Accordingly, the total dimension of the concatenated representation is $K_{\mathrm{tot}} \triangleq \sum_{m=1}^M K_m = MK$.

At each device, the raw input is first processed by the semantic encoder $f^m(\cdot;\boldsymbol{\alpha}^m)$ with parameters $\boldsymbol{\alpha}^m$, producing
\begin{equation}
\boldsymbol{z}^m = f^m\!\left(\boldsymbol{x}^m;\boldsymbol{\alpha}^m\right), \quad m\in\mathcal{M},
\end{equation}
where $\boldsymbol{z}^m$ denotes the semantic feature of modality $m$.
Each feature vector is then mapped into a sequence of complex channel symbols that occupies a predefined number of channel uses. For notational simplicity, we still denote the transmitted symbol vector by $\boldsymbol{z}^m$. The received signal at the server is
\begin{equation}
\hat{\boldsymbol{z}}^m = h^m\,\boldsymbol{z}^m + \boldsymbol{n}^m, \quad m\in\mathcal{M},
\end{equation}
where $h^m\!\in\!\mathbb{C}$ denotes the flat-fading channel coefficient for modality $m$, and
$\boldsymbol{n}^m\!\sim\!\mathcal{CN}\!\left(\boldsymbol{0},\,\sigma_{m}^{2}\boldsymbol{I}_{K}\right)$
represents additive white Gaussian noise (AWGN) with power $\sigma_{m}^{2}$.

Let $C_{\text{train}}$ and $C_{\text{infer}}$ denote the training-phase and inference-phase communication costs, respectively. They are quantified by the cumulative number of complex symbols (equivalently, communicated feature dimensions) exchanged between all devices and the server.
The system is subject to average communication budgets $\bar{C}_{\text{train}}$ and $\bar{C}_{\text{infer}}$, which capture the physical limits imposed by finite time--frequency resources and power constraints.

To balance communication efficiency and inference reliability, we incorporate an uncertainty-guided retransmission strategy $\pi$.
Specifically, the server first performs a provisional inference based on the currently received features and estimates the corresponding epistemic uncertainty.
If the uncertainty exceeds a predefined threshold, $\pi$ triggers a lightweight feedback request to selected devices for retransmission or additional feature dimensions.
Upon receiving the extra transmissions, the server refines the modality-wise received features by combining/augmenting all available observations; for notational simplicity, we still denote the resulting (post-retransmission) features by $\hat{\boldsymbol{z}}^{m}$.

At the server, the received modality-wise features are concatenated into a unified representation $\hat{\boldsymbol{z}} \in\mathbb{C}^{K_{\mathrm{tot}}}$:
\begin{equation}
\hat{\boldsymbol{z}} \triangleq \mathrm{concat}\!\big(\hat{\boldsymbol{z}}^{1}, \hat{\boldsymbol{z}}^{2}, \ldots, \hat{\boldsymbol{z}}^{M}\big)\in\mathbb{C}^{K_{\mathrm{tot}}}.
\end{equation}
Based on $\hat{\boldsymbol{z}}$, the central server performs semantic inference via a decoding function
\begin{equation}
\hat{\boldsymbol{y}} = g(\hat{\boldsymbol{z}};\boldsymbol{\phi}), \qquad 
g:\mathbb{C}^{K_{\mathrm{tot}}}\rightarrow \Delta^{C-1},
\end{equation}
where $g(\cdot;\boldsymbol{\phi})$ denotes the server-side decoder parameterized by $\boldsymbol{\phi}$.
The set $\Delta^{C-1}$ represents the $(C-1)$-dimensional probability simplex, i.e.,
\[
\Delta^{C-1}
\triangleq
\Big\{
\boldsymbol{p}\in\mathbb{R}^{C} \,\big|\, p_k\ge 0,\; \sum_{k=1}^{C} p_k = 1
\Big\},
\]
which characterizes all valid class-probability vectors over the $C$-class label space $\mathcal{Y}=\{1,\ldots,C\}$. 
The final predicted label is determined by the maximum a posteriori (MAP) rule, $\hat{y}=\arg\max_{k\in\mathcal{Y}}\hat{y}_{k}$.

\subsection{Problem Formulation}
In this paper, we aim to jointly design the modality-specific encoders $\{f^m\}$, the server-side decoder $g$, and the uncertainty-guided retransmission strategy $\pi$ to reduce both training- and inference-phase communication while maintaining robust performance under varying wireless channels and noisy modalities.
Specifically, we minimize the expected inference loss $\mathcal{L}(\hat{\boldsymbol{y}}, \boldsymbol{y})$ (e.g., cross-entropy loss) over the randomness of data samples, channel realizations, and noise, while keeping the average communication costs in both phases below given budgets. Formally, we consider:
\begin{align}
\min_{\{f^m\},\, g,\, \pi} \quad & \mathbb{E}\!\left[\mathcal{L}(\hat{\boldsymbol{y}}, \boldsymbol{y})\right] \label{eq:main_obj}\\
\text{s.t.} \quad & \mathbb{E}\!\left[C_{\text{train}}(\{f^m\}, g)\right] \leq \bar{C}_{\text{train}}, \label{eq:ctrain_constraint}\\
& \mathbb{E}\!\left[C_{\text{infer}}(\{f^m\}, g, \pi)\right] \leq \bar{C}_{\text{infer}}. \label{eq:cinfer_constraint}
\end{align}
This formulation induces a distributed optimization problem in which the encoders $\{f^m\}$ are optimized locally at devices with minimal device--server exchange. The decoder $g$ is learned at the server, and the inference-time policy $\pi$ is executed at the server to trigger selective retransmissions from devices when needed.

To operationalize this objective, we evaluate the proposed framework using the following key performance metrics:
\begin{itemize}
    \item \textbf{Classification Accuracy}: The proportion of correctly predicted labels, evaluated as $\frac{1}{N} \sum_{i=1}^N \mathbb{I}(\hat{y}_i = y_i)$, where $N$ is the number of samples and $\mathbb{I}(\cdot)$ is the indicator function. This serves as an empirical counterpart of the inference loss.
    \item \textbf{Robustness under Varying SNR}: Accuracy maintained across a range of SNR values, reflecting how well the inference loss can be controlled under different channel conditions.
    \item \textbf{Communication Efficiency}: Measured by the communication cost incurred during both training and inference, including (i) the number of device--server communication rounds required for training and (ii) the average amount of transmitted semantic features (including retransmissions) per sample during inference. These quantities correspond to $C_{\text{train}}$ and $C_{\text{infer}}$ in \eqref{eq:ctrain_constraint}--\eqref{eq:cinfer_constraint}.
\end{itemize}

These metrics align with the constrained objective in \eqref{eq:main_obj}--\eqref{eq:cinfer_constraint}. Accordingly, we next introduce a three-stage, communication-aware framework that yields a practical approximation to this coupled design under both training- and inference-phase budgets.

\section{Evidential Self-Supervised Framework for Distributed Multi-Modal Semantic Edge Inference}

\begin{figure*}[t]
  \centering
  \includegraphics[width=0.65\textwidth]{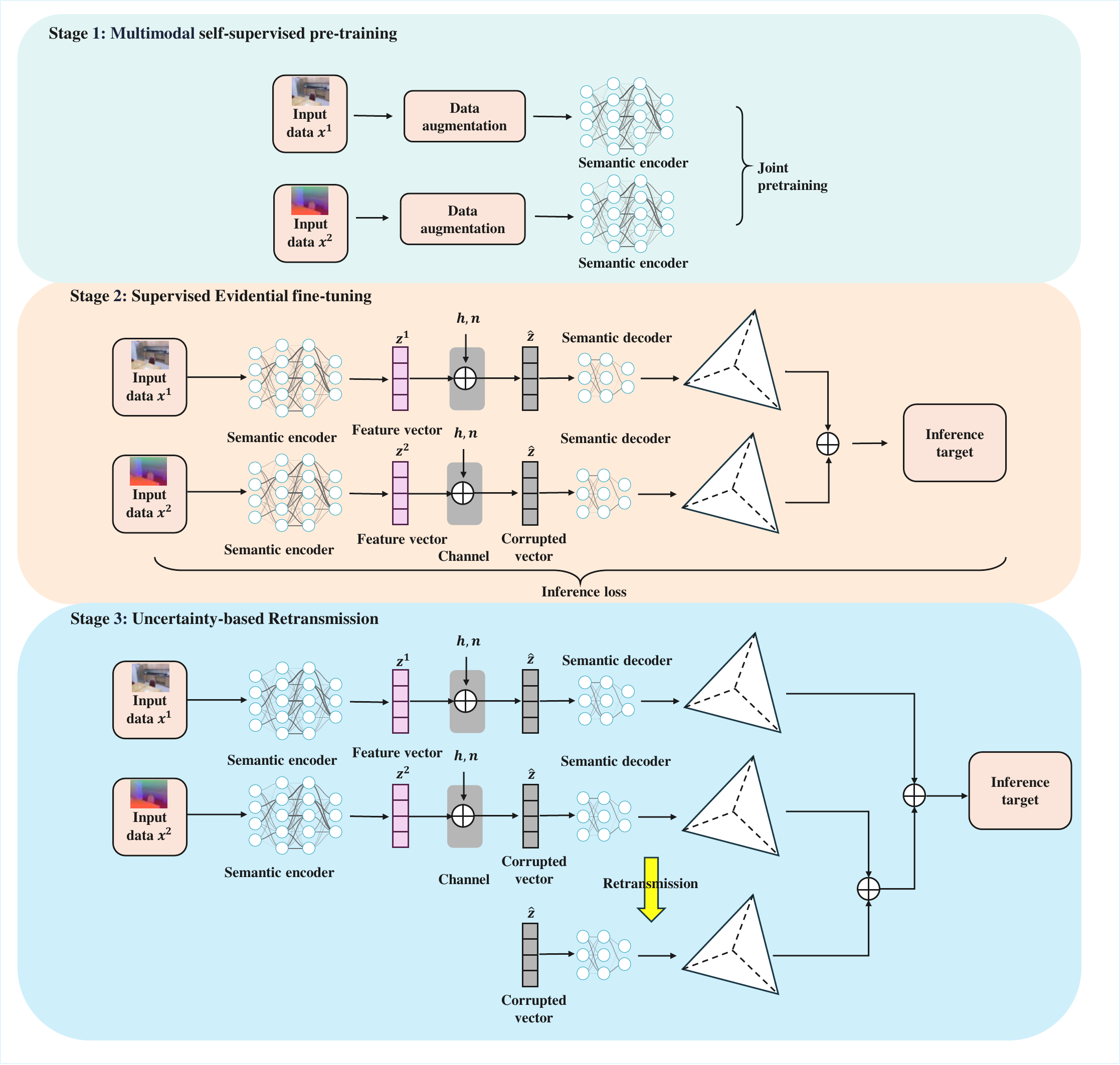}
  \caption{Three-stage evidential self-supervised \emph{multi-modal} semantic communication: (I) multi-modal self-supervised pre-training (no device–server communication), (II) uncertainty-aware supervised fine-tuning with reliable fusion, and (III) uncertainty-guided adaptive retransmission. Per-view heads output Dirichlet (evidential) distributions. Example shown for RGB and depth.}
  \label{fig:system}
\end{figure*}

In this section, we develop a three-stage evidential self-supervised framework that operationalizes the constrained optimization problem in \eqref{eq:main_obj}--\eqref{eq:cinfer_constraint}. The key idea is to decompose the coupled design. In particular, we first obtain a communication-free initialization that reduces the required supervised synchronization rounds (thereby minimizing $C_{\mathrm{train}}$). Then improve robustness and uncertainty calibration at fixed payload via evidential fusion (thereby reducing $\mathbb{E}[L(\hat y,y)]$). Finally, we enforce the inference budget by an uncertainty-guided retransmission policy that directly controls $C_{\mathrm{infer}}$. We begin by establishing an information-theoretic rationale for multi-modal self-supervised learning under both data augmentations and wireless distortions and explains why it improves communication efficiency. This serves as the motivation for design in Stage I.

\begin{figure}[htbp]
    \centering
    \includegraphics[width=2.6in]{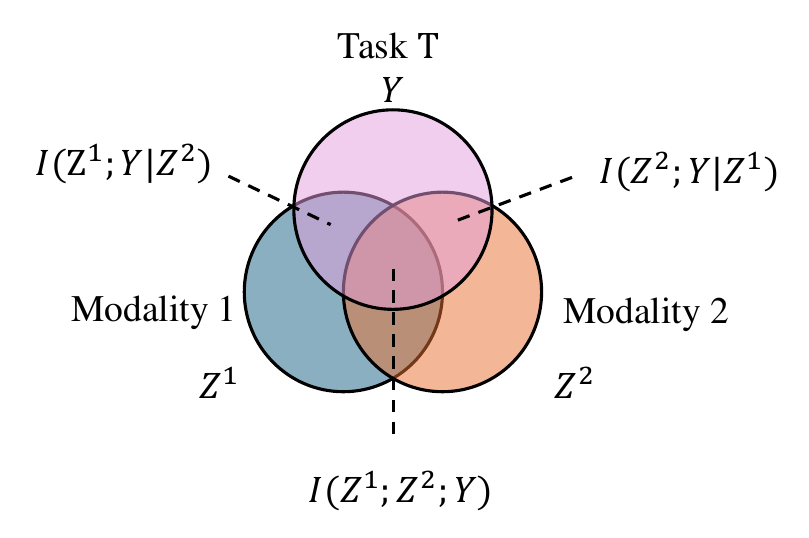}
    \caption{Mutual information analysis of single-modal versus multi-modal pre-training.}
    \label{venne}
\end{figure}
\subsection{Information-Theoretic Basis of Multi-Modal Self-Supervision}
\subsubsection{Intuitive Understanding of Multi-modal Pre-training}
For the two-modality case, the Venn diagram in Fig.~\ref{venne} provides an intuitive picture of how multi-modal pre-training can benefit downstream tasks.
Let $\boldsymbol{Z}^1$, $\boldsymbol{Z}^2$, and $\boldsymbol{Y}$ denote the random variables corresponding to the semantic features $\boldsymbol{z}^1$, $\boldsymbol{z}^2$ and the label $\boldsymbol{y}$, respectively.
The diagram highlights three key components:
(i) the shared cross-modal term $I(\boldsymbol{Z}^1;\boldsymbol{Z}^2;\boldsymbol{Y})$,
which denotes the interaction mutual information, (ii) the information that is unique to modality~1, $I(\boldsymbol{Z}^1;\boldsymbol{Y}\mid\boldsymbol{Z}^2)$,
and (iii) the information that is unique to modality~2, $I(\boldsymbol{Z}^2;\boldsymbol{Y}\mid\boldsymbol{Z}^1)$.
Here, the shared cross-modal term $I(\boldsymbol{Z}^1;\boldsymbol{Z}^2;\boldsymbol{Y})$ captures the label-relevant information that is redundantly shared by the two modalities. By the chain rule of mutual information, we decompose $I(\boldsymbol{Z}^1;\boldsymbol{Z}^2;\boldsymbol{Y})$ into two parts:
\begin{align}
I(\boldsymbol{Z}^1;\boldsymbol{Z}^2;\boldsymbol{Y})
&\triangleq I(\boldsymbol{Z}^1;\boldsymbol{Y}) - I(\boldsymbol{Z}^1;\boldsymbol{Y}\mid \boldsymbol{Z}^2) \notag\\
&= I(\boldsymbol{Z}^2;\boldsymbol{Y}) - I(\boldsymbol{Z}^2;\boldsymbol{Y}\mid \boldsymbol{Z}^1). \label{eq:interaction_mi_zy}
\end{align}

Classical single-modal pre-training methods\cite{chen2020simple, zbontar2021barlow} optimize one view at a time, focusing on $I(\boldsymbol{Z}^1;\boldsymbol{Y})$ or $I(\boldsymbol{Z}^2;\boldsymbol{Y})$.
When such objectives are extended to multi-modal settings, the learning dynamics tend to overweight the shared term $I(\boldsymbol{Z}^1;\boldsymbol{Z}^2;\boldsymbol{Y})$ and under-exploit the modality-specific terms $I(\boldsymbol{Z}^1;\boldsymbol{Y}\mid\boldsymbol{Z}^2)$ and $I(\boldsymbol{Z}^2;\boldsymbol{Y}\mid\boldsymbol{Z}^1)$, leading to a loss of complementary cross-modal cues. To preserve both the shared and modality-specific components, our multi-modal pre-training jointly optimizes the pair $(\boldsymbol{Z}^1,\boldsymbol{Z}^2)$

\subsubsection{Theoretical Guarantees for Multi-modal Pre-training}
We now formalize the intuition for $M$ modalities. 
Let $\boldsymbol{Z}^{1:M}=(\boldsymbol{Z}^{1},\ldots,\boldsymbol{Z}^{M})$ denote the underlying task-relevant signal associated with the $M$ observed views $\boldsymbol{X}^{1:M}=(\boldsymbol{X}^{1},\ldots,\boldsymbol{X}^{M})$. 
At each device $m$, a semantic feature $\boldsymbol{Z}^m$ is extracted as the latent signal of the corresponding view $\boldsymbol{X}^m$. 
Denote the received features at the server by $\hat{\boldsymbol{Z}} \!=\! \big(\hat{\boldsymbol{Z}}^{1},\ldots,\hat{\boldsymbol{Z}}^{M}\big)$. 
Each latent signal passes through an independent wireless channel, incurring an independent information loss defined as $\varepsilon_c^{m} \!\triangleq\! I(\boldsymbol{Z}^{m};\boldsymbol{X}^{m})-I(\hat{\boldsymbol{Z}}^{m};\boldsymbol{X}^{m})$, where the total communication loss is $\varepsilon_c \!=\! \sum_{m=1}^M \varepsilon_c^{m}$. 
The following result extends Theorem~1 of \cite{li2025remote} to the multi-modal setting:

\medskip
\begin{theorem}[Information-theoretic Guarantee for Multi-modal SSL]
\label{thm:mv-extension}
Consider an arbitrary subset of modalities $\mathcal{S} \subseteq \mathcal{M}$. All variables satisfy the probabilistic graphical model
\begin{equation}
X_{\mathcal{S}}' \leftarrow Y \rightarrow X_{\mathcal{S}} \rightarrow Z_{\mathcal{S}} \rightarrow \widehat{Z}_{\mathcal{S}},
\label{eq:mv-pgm}
\end{equation}
where $X_{\mathcal{S}}'$ denote augmented views of the same modalities (i.e., data samples transformed via operations like random cropping or color jittering that preserve the underlying semantic information $Y$) and $\widehat{Z}_{\mathcal{S}}$ are the representations received after the noisy wireless channel (channel-induced information loss $\varepsilon_{\mathrm{c}}(\mathcal{S}) \triangleq I(Z_{\mathcal{S}}; X_{\mathcal{S}}') - I(\widehat{Z}_{\mathcal{S}}; X_{\mathcal{S}}') \ge 0$).


Under perfect channels ($\varepsilon_{\mathrm{c}}(\mathcal{S})=0$) and supervised learning, the minimal sufficient representation preserves all task-relevant information:
\begin{equation} 
I(\widehat{Z}_{\mathcal{S},\mathrm{sup}}^{\mathrm{opt,min}}; Y) = I(X_{\mathcal{S}}; Y).
\end{equation}
In the practical self-supervised setting with channel noise, the corresponding minimal representation learned by multi-modal SSL satisfies
\begin{equation}
\begin{aligned}
I(X_{\mathcal{S}}; Y) &\ge I(\widehat{Z}_{\mathcal{S},\mathrm{ssl}}^{\mathrm{opt,min}}; Y) \\
&\ge I(X_{\mathcal{S}}; Y) - \underbrace{I(X_{\mathcal{S}}; Y \mid X_{\mathcal{S}}')}_{\text{augmentation gap}} - \underbrace{\varepsilon_{\mathrm{c}}(\mathcal{S})}_{\text{channel loss}}.
\end{aligned}
\label{eq:mv-bound-main}
\end{equation}
This indicates that its task-relevant information is rigorously sandwiched between the supervised optimum and a quantifiable lower bound depending only on augmentation strength and channel quality.

\emph{The proof is given in the Appendix~\ref{app:proof-mv}.} 
\end{theorem}

\paragraph*{Discussion}
Theorem~\ref{thm:mv-extension} implies that, under the probabilistic model in~\eqref{eq:mv-pgm}, multi-modal self-supervised pre-training can recover representations whose label information $I(\widehat{Z}_{\mathcal{S},\mathrm{ssl}}^{\mathrm{opt,min}}; Y)$ approaches that of the supervised optimum, up to the irreducible losses induced by augmentations and wireless channels. Consequently, achieving a desired task performance may require substantially less labeled supervision in the subsequent adaptation stage.
For distributed edge inference, this provides a direct communication benefit: since device--server communication is only needed during supervised fine-tuning, reducing the number of fine-tuning epochs lowers the training-phase communication cost $C_{\text{train}}$ in Section~II.

\subsection{Evidential Multi-Modal Semantic Communication Architecture}

\subsubsection{Stage I: Multi-Modal Pre-Training}
Recent studies such as~\cite{wang2023decur} show that disentangling shared and modality-specific representations is critical for integrating complementary multi-modal evidence. Motivated by this observation and by the communication constraints in distributed edge training, we design a self-supervised pre-training objective that (i) explicitly decouples cross-modal shared factors from modality-specific factors, and (ii) can be executed fully locally at each device without transmitting raw data, features, or gradients. As a result, Stage~I provides a \emph{communication-free initialization} for subsequent supervised adaptation, effectively reducing $C_{\text{train}}$ by eliminating device--server exchanges during pre-training and improving distributed convergence efficiency.

\begin{equation}
    \mathcal{L}_{\text{pre-train}} = \mathcal{L}_\text{cross} + \mathcal{L}_\text{intra},
    \label{eq:pretrain_loss}
\end{equation}
where \( \mathcal{L}_\text{cross} \) captures cross-modal consistency and \( \mathcal{L}_\text{intra} \) focuses on intra-modal robustness.  

Let $\mathcal{D}=\{(\boldsymbol{x}^{1:M}_i,y_i)\}_{i=1}^N$ denote the training set. Stage~I performs self-supervised learning using only the multi-modal inputs $\boldsymbol{x}^{1:M}$.
For each modality $m$, we generate an augmented view $\tilde{\boldsymbol{x}}^{m}$ from $\boldsymbol{x}^{m}$ through a randomized augmentation operator, and encode the paired views to obtain
$\{\boldsymbol{z}^m,\tilde{\boldsymbol{z}}^m\}_{m=1}^M$.
All losses below are computed on mini-batches of size $B$.

For each modality $m$, we compute a batch-wise normalized cross-correlation matrix $\mathcal{C}^m\in\mathbb{R}^{K\times K}$ between the two augmented features:
\begin{equation}
\mathcal{C}^{m}_{ij}
=
\frac{\sum_{b=1}^B z_{b,i}^m \tilde{z}_{b,j}^m}
{\sqrt{\sum_{b=1}^B (z_{b,i}^m)^2}\sqrt{\sum_{b=1}^B (\tilde{z}_{b,j}^m)^2}},
\end{equation}
where $i,j\in\{1,\ldots,K\}$ index feature dimensions. The diagonal entries $\mathcal{C}^m_{ii}$ measure same-dimension alignment across augmented views (augmentation invariance), whereas off-diagonal entries quantify cross-dimension redundancy.

The intra-modal loss is then defined as:
\begin{align}
    \mathcal{L}_{\text{intra}} &= \sum_{m=1}^M \mathcal{L}_{\text{intra}}^{m}, \\
    \mathcal{L}_{\text{intra}}^{m} &= \underbrace{\sum_{i=1}^K \left(1 - \mathcal{C}^{m}_{ii}\right)^2}_{\text{Diagonal alignment}} 
    + \lambda_m \cdot \underbrace{\sum_{i \neq j} \left( \mathcal{C}^{m}_{ij} \right)^2}_{\text{Off-diagonal decorrelation}},
\end{align}
where $\lambda_m > 0$ balances feature stability (diagonal) and redundancy reduction (off-diagonal).

For cross-modal interactions, we compute the correlation matrix $\mathcal{C}^{m,n}$ between modalities $m$ and $n$:
\begin{equation}
    \mathcal{C}^{m,n}_{ij} = \frac{\sum_{b=1}^B z_{b,i}^m z_{b,j}^n}{\sqrt{\sum_{b=1}^B ( z_{b,i}^m )^2} \sqrt{\sum_{b=1}^B ( z_{b,j}^n )^2}},
\end{equation}
which captures pairwise correlations between cross-modal feature dimensions.

To disentangle shared and unique representations, $\mathcal{C}^{m,n}$ is partitioned into:
\begin{itemize}
    \item \textbf{Shared Representation Block} ($\mathcal{C}_\text{sha}$): captures modality-invariant features.  
    \item \textbf{Unique Representation Block} ($\mathcal{C}_\text{uni}$): retains modality-specific information.  
\end{itemize}
Here $K_\text{sha}+K_\text{uni}=K$ ensures dimensional consistency.

The cross-modal loss is then formulated as:
\begin{equation}
    \mathcal{L}_\text{cross} = \sum_{m=1}^M \sum_{n \neq m}^N \left( \mathcal{L}_\text{sha}^{m,n} + \mathcal{L}_\text{uni}^{m,n} \right),
\end{equation}
where:
\begin{alignat}{3}
    \mathcal{L}_\text{sha}^{m,n} &= \sum_{i=1}^{K_\text{sha}} \left(1 - \mathcal{C}_{\text{sha},ii}^{m,n}\right)^2 
    + \lambda_\text{sha} \sum_{i \neq j} \left(\mathcal{C}_{\text{sha},ij}^{m,n}\right)^2, \\
    \mathcal{L}_\text{uni}^{m,n} &= \sum_{i=1}^{K_\text{uni}} \left(\mathcal{C}_{\text{uni},ii}^{m,n}\right)^2 
    + \lambda_\text{uni} \sum_{i \neq j} \left(\mathcal{C}_{\text{uni},ij}^{m,n}\right)^2,
\end{alignat}
with $\lambda_\text{sha}, \lambda_\text{uni} > 0$ controlling the balance between shared alignment and unique disentanglement.

Finally, the pre-trained encoders are obtained as:
\begin{equation}
\left\{\boldsymbol{\alpha}^{m\dagger}\right\}_{m=1}^M = \arg\min_{ \{\boldsymbol{\alpha}^{m}\}_{m=1}^M} \mathcal{L}_{\text{pre-train}}.
\end{equation}

Since all operations in Stage~I are performed locally at the edge devices without exchanging raw data, features, or gradients with the server, this pre-training stage does not contribute to the device--server communication cost $C_{\text{train}}$ and effectively provides a communication-free initialization for subsequent supervised adaptation.

\subsubsection{Stage II: Uncertainty-Aware Supervised Fine-Tuning}

In the second stage, the pre-trained encoders residing on the edge devices and the decoder deployed at the central server are further refined using supervised labels. Stage II serves as a centralized global refinement step that aggregates modality-wise representations obtained from distributed devices, improving robustness under distributed sensing noise. Unlike a standard cross-entropy baseline, we adopt an evidential deep learning formulation~\cite{han2022trusted}. This framework explicitly models predictive uncertainty and exploits it to enhance robustness in varying wireless environments. Evidential heads and fusion are applied on the server side. The fine-tuning procedure consists of three key modules: (i) \emph{modality-specific} evidential learning heads on top of encoders, (ii) uncertainty-aware \emph{fusion across modalities}, and (iii) the overall optimization objective. 

\paragraph{Modality-Specific Evidential Deep Learning}

For a sample $\boldsymbol{x}_i^{\,m}$ from modality $m$, the encoder produces a semantic feature
\begin{equation}
\boldsymbol{z}_i^{\,m} = f^m(\boldsymbol{x}_i^{\,m};\boldsymbol{\alpha}^m),
\end{equation}
which is conveyed over the wireless channel to yield the received feature $\hat{\boldsymbol{z}}_i^{\,m}$. An evidential head
$h^m(\cdot;\boldsymbol{\psi}^m)$ maps the (possibly noisy) received feature to evidence logits, followed by a non-negative activation $a(\cdot)$ (e.g., ReLU):
\begin{equation}
\boldsymbol{e}_i^{\,m} = a\!\big(h^m(\hat{\boldsymbol{z}}_i^{\,m};\boldsymbol{\psi}^m)\big)
= [\,e_{i1}^{\,m},\ldots,e_{iC}^{\,m}\,],
\end{equation}
where $C$ denotes the number of classes. The evidence induces a Dirichlet distribution with concentration parameters
\begin{equation}
\boldsymbol{\gamma}_i^{\,m} = \boldsymbol{e}_i^{\,m} + \mathbf{1}.
\end{equation}
From $\operatorname{Dir}(\boldsymbol{\gamma}_i^{\,m})$, the belief mass for class $c$ and the epistemic uncertainty are
\begin{equation}
b_{ic}^{\,m} = \frac{e_{ic}^{\,m}}{S_i^{\,m}},\quad
u_i^{\,m} = \frac{C}{S_i^{\,m}} = 1-\sum_{c=1}^{C} b_{ic}^{\,m},\quad
S_i^{\,m} = \sum_{c=1}^{C} e_{ic}^{\,m} + C .
\end{equation}

\paragraph{Uncertainty-Aware Fusion Across Modalities}

Each modality produces an opinion $\mathbb{M}_i^{\,m}=\{\boldsymbol{b}_i^{\,m},\,u_i^{\,m}\}$. For two modalities (e.g., $m=1,2$), following~\cite{han2022trusted}, the fusion operator $\oplus$ is defined componentwise as

\begin{equation}
\begin{aligned}
(\boldsymbol{b}_i^{\,1} \oplus \boldsymbol{b}_i^{\,2})_c
&= \frac{b_{ic}^{\,1} \, u_i^{\,2} + b_{ic}^{\,2} \, u_i^{\,1}}
         {u_i^{\,1} + u_i^{\,2} - u_i^{\,1}u_i^{\,2}}, \\[8pt]
u_i^{\,1} \oplus u_i^{\,2}
&= \frac{u_i^{\,1} u_i^{\,2}}
         {u_i^{\,1} + u_i^{\,2} - u_i^{\,1}u_i^{\,2}}.
\end{aligned}
\end{equation}
By associativity, this extends to $M$ modalities:
\[
\mathbb{M}_i = \bigoplus_{m=1}^{M} \mathbb{M}_i^{\,m}.
\]

\paragraph{Overall Training Objective}

We optimize the encoders $\{f^m(\cdot;\boldsymbol{\alpha}^m)\}_{m=1}^M$, evidential heads $\{h^m(\cdot;\boldsymbol{\psi}^m)\}_{m=1}^M$, and the decoder $g(\cdot;\boldsymbol{\phi})$ jointly. Let $\Theta=\{\boldsymbol{\alpha}^m,\boldsymbol{\psi}^m\}_{m=1}^M \cup \{\boldsymbol{\phi}\}$ denote all trainable parameters.

The accuracy term is the negative log-likelihood under the Dirichlet predictive distribution:
\begin{equation}
\begin{aligned}
\ell_{\text{acc}}(\Theta)_i &= \mathbb{E}_{\boldsymbol{p}_i \sim \operatorname{Dir}(\boldsymbol{p}_i \mid \boldsymbol{\gamma}_i)}
\!\left[-\sum_{c=1}^{C} y_{ic}\log(p_{ic}) \right] \\
&= \sum_{c=1}^{C} y_{ic} \left( \log(S_i) - \log(e_{ic}+1) \right),
\end{aligned}
\end{equation}
or equivalently in digamma form,
\begin{equation}
\ell_{\text{acc}}(\Theta)_i
= \sum_{c=1}^{C} y_{ic}\Big(\psi(S_i)-\psi(\gamma_{ic})\Big),
\end{equation}
where $S_i=\sum_{c=1}^{C} e_{ic}+C$, $\gamma_{ic}=e_{ic}+1$, and $\psi(\cdot)$ is the digamma function. (When using modality-wise terms, replace $e_{ic},S_i$ with $e_{ic}^{\,m},S_i^{\,m}$.)

To suppress spurious evidence, we add a KL regularizer toward the uniform Dirichlet prior:
\begin{equation}
\begin{aligned}
\ell_{\mathrm{KL}}(\Theta)_i &= \lambda_t\, \mathrm{KL}\!\left[ \operatorname{Dir}(\tilde{\boldsymbol{\gamma}}_i)\, \|\, \operatorname{Dir}(\mathbf{1}) \right], \\
\tilde{\boldsymbol{\gamma}}_i &= \boldsymbol{y}_i + (1-\boldsymbol{y}_i)\odot \boldsymbol{\gamma}_i
\end{aligned}
\end{equation}
with annealing
\begin{equation}
\lambda_t = \lambda_0 \exp\!\left(-\frac{\ln\lambda_0}{T}t\right),
\end{equation}
where $t$ is the current epoch, $T$ the total epochs, and $\lambda_0$ a small constant.

The final fine-tuning objective combines the accuracy and KL terms as
\begin{equation}
\mathcal{L}_{\text{fine-tuning}}(\Theta)
= \sum_{i=1}^{n} \left( \ell_{\text{acc}}(\Theta)_i + \ell_{\mathrm{KL}}(\Theta)_i \right),
\label{eq:finetune_loss}
\end{equation}
which is minimized jointly over the model parameters $\Theta$. This yields accurate predictions with well-calibrated uncertainty and robustness, which will be exploited by the retransmission policy in Stage~III. Since all evidential computations are performed at the server using the transmitted semantic features, Stage~II enhances robustness without incurring any additional device–server communication cost. Consequently, it increases the \emph{task utility per transmitted bit} by extracting more reliable decisions from these features.

\subsubsection{Stage III: Adaptive Retransmission Mechanism}
Stage III implements an uncertainty-guided feedback policy where the server computes per-modality epistemic uncertainty from evidential heads and requests retransmissions from selected devices when the uncertainty exceeds a threshold. To address this issue under explicit communication constraints, we propose an adaptive retransmission mechanism that leverages \emph{per-modality} predictive uncertainty. This mechanism can be viewed as an inference-time communication policy $\pi$ that decides, for each modality and each sample, whether additional transmissions are necessary, thereby trading a controllable amount of extra communication for improved reliability. Compared with prior works~\cite{gao2023adaptive,shao2022task}, our approach achieves higher efficiency without introducing additional modules. Specifically, for modality $m$, the retransmission decision is made based on the following criterion:
\begin{equation}
r(\hat{\boldsymbol{z}}_i^{\,m}) =
\begin{cases}
1, & \text{if } u(\hat{\boldsymbol{z}}_i^{\,m}) \ge u_\lambda,\\
0, & \text{otherwise},
\end{cases}
\end{equation}
where $u(\hat{\boldsymbol{z}}_i^{\,m})$ denotes the estimated uncertainty for input $\hat{\boldsymbol{z}}_i^{\,m}$ and $u_\lambda$ is a learnable threshold determined directly on the training set.

When $r(\hat{\boldsymbol{z}}_i^{\,m})=1$, the sample of modality $m$ is retransmitted, enabling additional feature observations and reducing error probability. In this way, retransmission resources are allocated selectively, focusing on uncertain instances within each modality and improving communication efficiency.

For retransmitted samples, we first perform \emph{intra-modality} fusion to combine multiple observations of modality $m$, producing a fused opinion $\mathbb{M}_m^{f} = \{\boldsymbol{b}_m^{f}, u_m^{f}\}$. The updated uncertainty $u_m^{f}$ is then checked: if $u_m^{f} \geq u_\lambda$ and the number of retransmissions has not exceeded the maximum limit $N_{\max}$, the process continues for that modality. Otherwise, when $u_m^{f} < u_\lambda$, the system proceeds to \emph{cross-modality} fusion using the rule described in Stage~II.

To obtain a reliable threshold $u_\lambda$ without requiring a separate calibration or validation set, we directly optimize it on the training set using a quantile-based objective. Specifically, we choose $u_\lambda$ as the $(1-\alpha)$-quantile of the predictive uncertainty distribution over all correctly classified training samples:
\begin{equation}
u_\lambda = \text{quantile}\left( \left\{ u(\boldsymbol{z}_i^{\,m}) \mid i \in \mathcal{D}_{\text{train}}, \, \hat{y}_i = y_i \right\}, 1-\alpha \right),
\end{equation}
where $\alpha \in (0,1)$ is a hyperparameter that controls the desired retransmission rate (typically set to 0.2 in our experiments). This strategy ensures that retransmission is triggered only for the top-$\alpha$ most uncertain modalities among training samples that were already correctly classified under ideal conditions, thereby providing a principled way to control the average retransmission overhead while maintaining high final accuracy.


This training-set quantile threshold achieves similar statistical guarantees to validation-set calibration methods in practice, while being simpler to implement and fully compatible with end-to-end training pipelines.

\subsection{Three-Stage Training and Inference Procedure}

We now integrate the three stages into a unified training framework. In \textbf{Stage I}, multi-modal self-supervised pre-training is performed to extract both shared and modality-specific features, yielding task-agnostic representations. In \textbf{Stage II}, these pre-trained encoders $\{f^m(\cdot; \boldsymbol{\alpha}^m)\}_{m=1}^M$ are fine-tuned together with the decoder $g(\cdot;\boldsymbol{\phi})$ under supervised guidance, where evidential deep learning and reliable \emph{cross-modality} fusion ensure calibrated predictions. Finally, \textbf{Stage III} introduces an adaptive retransmission mechanism that leverages \emph{per-modality} predictive uncertainty to improve robustness against channel variations while minimizing redundant communication.  

The overall procedure results in a universal yet task-adaptive model that jointly balances efficiency, reliability, and robustness in multi-modal semantic communication. From a system perspective, the three stages together provide a communication-aware solution to the constrained optimization problem in Section II: Stage~I reduces $C_{\text{train}}$ via zero-communication distributed pre-training, Stage~II improves robustness and task utility per transmitted bit at fixed communication cost, and Stage~III adaptively controls $C_{\text{infer}}$ through uncertainty-guided retransmissions. The detailed training process of the proposed algorithm is summarized in Algorithm~\ref{alg:self_supervised}.


\begin{algorithm}[t]
\caption{Robust Uncertainty-Based Multi-modal Semantic Communication}
\label{alg:self_supervised}
\begin{algorithmic}[1]
\STATE \textbf{Input:} training set $\mathcal{D}$, batch size $B$, hyperparameters $\boldsymbol{\lambda}$, epochs $E_1,E_2$, uncertainty threshold $u_\lambda$, max retransmissions $N_{\max}$.
\STATE \textbf{Output:} encoder parameters $\{f^m(\cdot;\boldsymbol{\alpha}^m)\}_{m=1}^M$ and decoder
$g(\cdot;\boldsymbol{\phi})$.

\STATE \textbf{Stage I: Multi-modal pre-training}
\STATE Initialize $\{\boldsymbol{\alpha}^m\}_{m=1}^M$.
\FOR{$e=1$ to $E_1$}
  \STATE Sample a minibatch from $\mathcal{D}$ and generate augmented views $\{\boldsymbol{x}^m,\tilde{\boldsymbol{x}}^m\}_{m=1}^M$.
  \STATE Compute pre-training loss $\mathcal{L}_{\text{pre-train}}$ in~\eqref{eq:pretrain_loss} and update $\{\boldsymbol{\alpha}^m\}_{m=1}^M$.
\ENDFOR

\STATE \textbf{Stage II: Supervised fine-tuning with evidential fusion}
\STATE Initialize $\{\boldsymbol{\alpha}^{m\dagger}\}_{m=1}^M \gets \{\boldsymbol{\alpha}^m\}_{m=1}^M$.
\FOR{$e=1$ to $E_2$}
  \STATE Sample a minibatch from $\mathcal{D}$ and obtain received features $\{\hat{\boldsymbol{z}}^m\}_{m=1}^M$.
  \STATE Compute supervised loss $\mathcal{L}_{\text{sup}}$ in~\eqref{eq:finetune_loss} and update $\{\boldsymbol{\alpha}^m\}_{m=1}^M$ and $\boldsymbol{\phi}$.
\ENDFOR

\STATE \textbf{Stage III: Inference with uncertainty-based retransmission}
\FOR{each test sample}
  \FOR{$m=1$ to $M$}
    \STATE Encode and transmit to obtain $\hat{\boldsymbol{z}}^{m,0}$ and opinion $O_m^0=(\boldsymbol{b}_m^0,u_m^0)$.
    \FOR{$n=1$ to $N_{\max}$}
      \IF{$u_m^{n-1} < u_\lambda$}
        \STATE \textbf{break}
      \ENDIF
      \STATE Retransmit to obtain $\hat{\boldsymbol{z}}^{m,n}$ and fuse with $O_m^{n-1}$ to get $O_m^{n}$.
    \ENDFOR
  \ENDFOR
  \STATE Fuse $\{O_m\}_{m=1}^M$ across modalities to obtain final opinion $O$ and prediction $\hat{\boldsymbol{y}} = g(O;\boldsymbol{\phi})$.
\ENDFOR
\end{algorithmic}
\end{algorithm}

\section{Numerical Results}
In this section, we evaluate the proposed multi-stage training framework for distributed multi-modal semantic communication, together with the uncertainty-guided retransmission mechanism. We consider edge-assisted image classification where devices transmit semantic features over bandwidth-limited wireless links to an edge/server for joint inference. We further study robustness under varying channel conditions and noisy inputs, and quantify training epochs and retransmission ratios as device--server communication cost.

\subsection{Experimental Setup}

\subsubsection{Datasets and Modalities}
We conduct experiments on two indoor RGB-D benchmarks: NYU Depth V2 (NYUDv2) and SUN RGB-D. NYUDv2 contains 1{,}449 RGB-D pairs from 27 scene categories~\cite{silberman2012indoor}; following~\cite{han2022trusted}, we merge them into 10 classes and use 795/654 images for training/testing. SUN RGB-D has 10{,}335 RGB-D images collected by four sensors~\cite{song2015sun}, providing larger scale and diversity to validate robustness and generalization. In all experiments, RGB and depth are treated as two modalities from two logical devices, each with its own encoder communicating with the edge/server as in Section~II.

\subsubsection{Model Architecture, Training Protocol, and Channel Settings}
RGB and depth encoders adopt ResNet-18~\cite{he2016deep} at the devices, while evidential heads and the decoder are placed at the edge/server. Models are trained on an NVIDIA A40 GPU with images resized to $256\times256$ and randomly cropped to $224\times224$ (batch size 64). Pre-training and fine-tuning run for 500 epochs (split in Section~III); in our synchronous implementation, each epoch equals one device--server communication round. Unless otherwise stated, links are modeled as AWGN channels and SNR is defined accordingly. We use SGD with weight decay and a cosine learning-rate schedule.

\subsubsection{Baselines and Ablation Variants}
We design baselines and ablations to attribute performance gains to the correct stage of the proposed pipeline. 
The framework addresses two coupled but mechanistically distinct objectives: reducing the training-phase device--server communication cost $C_{\mathrm{train}}$, and improving inference robustness under channel and sensing distortions with a controlled inference cost $C_{\mathrm{infer}}$. 
Stage~I mainly determines the quality and label-efficiency of the learned representations before any transmission, which directly affects the number of supervised fine-tuning rounds required and thus $C_{\mathrm{train}}$. 
Stage~II and Stage~III mainly determine how the received features are fused and when additional transmissions are requested, which affects reliability under varying channels/noisy modalities and thus the accuracy--$C_{\mathrm{infer}}$ tradeoff. 
Accordingly, we compare (i) encoder pre-training strategies for Stage~I and (ii) semantic communication schemes for Stage~II/III under matched architectures and channel settings. 
For Stage~I comparisons, we keep the same backbone and the same supervised fine-tuning protocol and disable retransmission to isolate representation learning effects. 
For Stage~II/III comparisons, we fix the encoder architecture and vary only the decision head and feedback policy to isolate the contributions of evidential fusion and uncertainty-guided retransmission.

\textbf{Stage~I: Pre-training strategies.}
Unless otherwise stated, all methods share the same backbone and supervised fine-tuning protocol at the server, and the retransmission mechanism is disabled.
\begin{itemize}
    \item \textbf{Proposed multi-modal pre-training}: Joint self-supervised pre-training that disentangles shared and modality-specific representations using the loss in~\eqref{eq:pretrain_loss}, followed by supervised fine-tuning at the server.
    \item \textbf{SimCLR}~\cite{chen2020simple}: Each modality is pre-trained independently with the SimCLR contrastive objective, then jointly fine-tuned with labels.
    \item \textbf{Barlow Twins}~\cite{zbontar2021barlow}: A redundancy-reduction baseline that maximizes cross-correlation between augmented views, followed by supervised fine-tuning.
    \item \textbf{Supervised (no pre-training)}: Encoders and decoder are trained from scratch using task labels only.
\end{itemize}

\textbf{Stage~II \& III: Semantic communication schemes.}
All schemes use the same encoders but differ in decision heads and feedback policy.
\begin{itemize}
    \item \textbf{JSCC baseline}~\cite{bourtsoulatze2019deep}: End-to-end task-oriented JSCC with self-supervised pre-training but without evidential modeling; fusion is performed by simple feature concatenation and there is no retransmission.
    \item \textbf{Evidential (no retransmission)}~\cite{han2022trusted}: Our framework with multi-modal pre-training and evidential reliability-aware fusion, but with one-shot transmission (no feedback) in Stage~III, isolating the effect of evidential modeling.
    \item \textbf{Evidential + retransmission}: The full method combining evidential learning with uncertainty-guided retransmissions in Stage~III. A retransmission is triggered when predictive uncertainty exceeds $u_\lambda$, with at most $N_{\max}{=}3$ attempts. Empirically, the average retransmission ratio is about $10\%$, so only a small fraction of samples require extra channel uses while robustness is improved.
\end{itemize}

\subsection{Training Communication Efficiency and Task Performance}
The proposed multi-modal self-supervised pre-training strategy is designed to learn label-efficient, noise-tolerant representations, thereby explicitly reducing the number of device--server communication rounds required during the training phase. We first employ controlled synthetic datasets to verify the effectiveness of this pre-training approach in preserving and disentangling shared and modality-specific factors as predicted by the theory. Subsequently, we evaluate its impact on training-phase communication efficiency using real-world indoor scene classification tasks.
\subsubsection{Representation Disentanglement on Synthetic Data}
Stage~I aims to disentangle shared cross-modal semantics and modality-specific information in $\{Z_1,Z_2\}$. We validate this using controlled synthetic data with known latent factors and probe mutual information.

\textit{a) Synthetic data generation:}
Let $d=32$ and sample $w_1,w_2,w_s \sim \mathcal{N}(0_d,\Sigma_d^2)$, where $w_s$ is shared and $w_1,w_2$ are modality-unique. We generate $x_1=T_1([w_1,w_s])$, $x_2=T_2([w_2,w_s])$, and produce $y$ from a nonlinear mixture of $w_s,w_1,w_2$ under AWGN to control shared/unique task-relevant signal.

\textit{b) Results:}
Tables~\ref{tab:mi_synth} and~\ref{tab:lp_snr} show that SimCLR/Barlow Twins mainly capture shared factors but little view-specific signal, whereas our multi-pretrained encoder preserves both shared and unique information, achieving the largest total informativeness ($I_{\text{wall}}=41.65$) and the best linear-probing accuracy across SNRs. These results suggest that Stage~I learns more disentangled and noise-robust multi-modal representations, and its SSL initialization substantially reduces distributed training communication.

\begin{table}[t]
\centering
\caption{Mutual-information probing on synthetic data (in bits).
$I_{w_1}:=I(Z_i; w_1)$, $I_{w_2}:=I(Z_i; w_2)$, $I_{w_s}:=I(Z_i; w_s)$.
The last row reports the total $I_{\text{wall}}$ across both views $(Z_1, Z_2)$ and all factors.}
\label{tab:mi_synth}
\begin{tabular}{l cc cc cc}
\hline
Metric & \multicolumn{2}{c}{SimCLR} & \multicolumn{2}{c}{Barlow Twins} & \multicolumn{2}{c}{Multi-pretrained (ours)} \\
       & $Z_1$ & $Z_2$ & $Z_1$ & $Z_2$ & $Z_1$ & $Z_2$ \\
\hline
$I_{w_1}$ & 8.68 & 0.43 & 1.77 & 0.44 & 12.01 & 0.68 \\
$I_{w_2}$ & 0.43 & 9.16 & 0.48 & 1.83 & 0.67 & 13.60 \\
$I_{w_s}$ & 12.86 & 11.47 & 15.83 & 15.81 & 14.45 & 13.74 \\
\hline
$I_{\text{wall}}$ & \multicolumn{2}{c}{27.79} & \multicolumn{2}{c}{19.16} & \multicolumn{2}{c}{41.65} \\
\hline
\end{tabular}
\end{table}

\begin{table}[t]
\centering
\caption{Linear-probing accuracy on synthetic data under different SNR levels.}
\label{tab:lp_snr}
\begin{tabular}{c c c c}
\hline
SNR (dB) & SimCLR & Barlow Twins & Proposed \\
\hline
0  & 0.744 & 0.691 & 0.748 \\
10 & 0.891 & 0.781 & 0.894 \\
20 & 0.957 & 0.874 & 0.962 \\
\hline
\end{tabular}
\end{table}

\subsubsection{Training-Phase Communication Efficiency}

\begin{figure*}[t]
    \centering
    \captionsetup[subfigure]{labelformat=empty,font=footnotesize}
    
    \subfloat[(a) Full labels, $\text{SNR}= 10$ dB]{
        \includegraphics[width=0.31\textwidth]{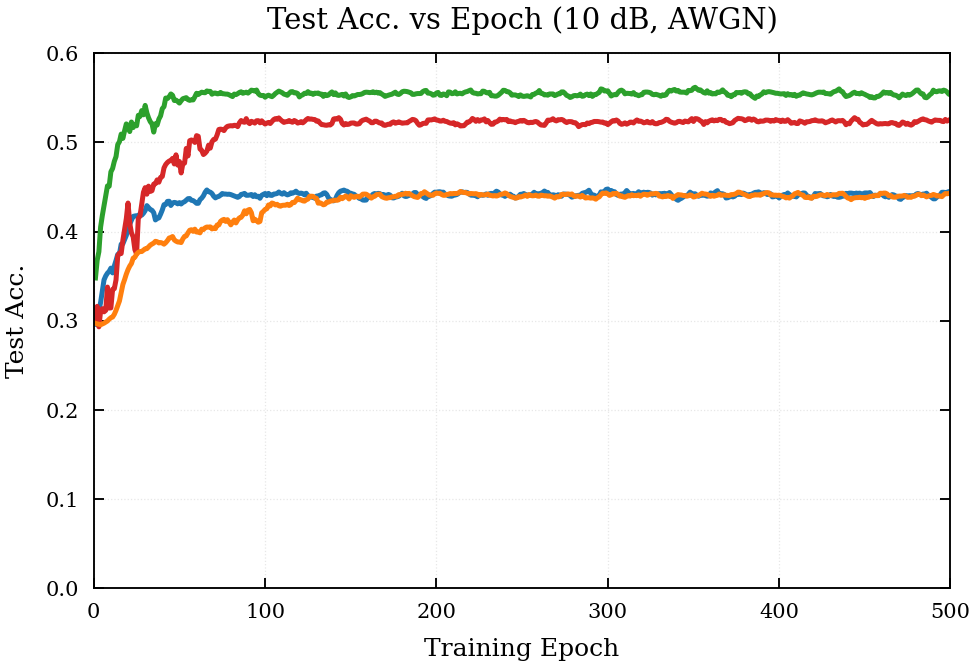}
    }\hspace{0.04\textwidth}
    \subfloat[(b) Full labels, $\text{SNR}= 20$ dB]{
        \includegraphics[width=0.31\textwidth]{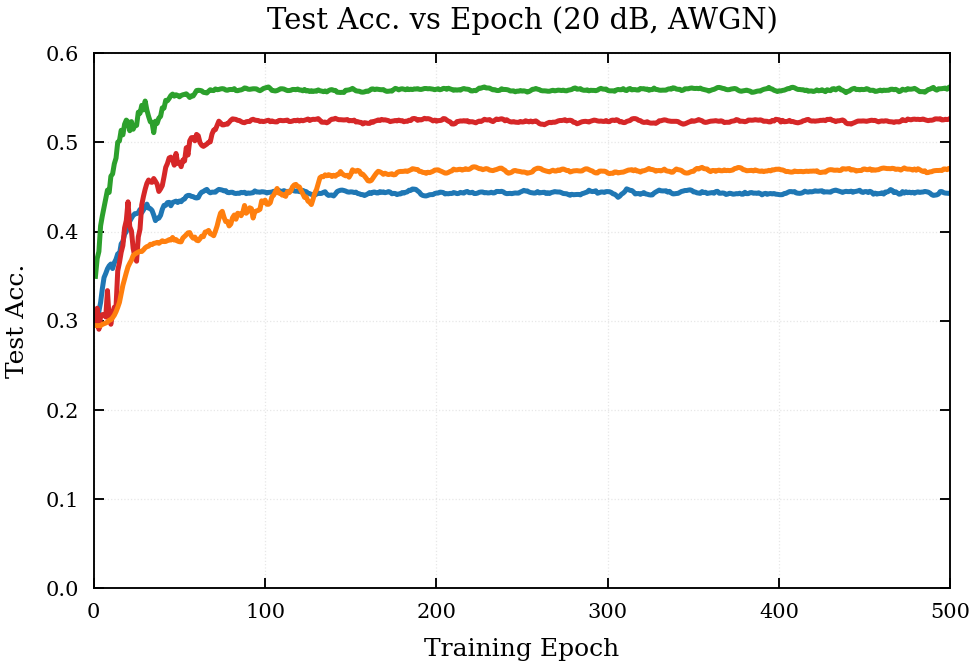}
    }

    \par\vspace{0.5ex}   

    \subfloat[(c) Few labels, $\text{SNR}= 10$ dB]{
        \includegraphics[width=0.31\textwidth]{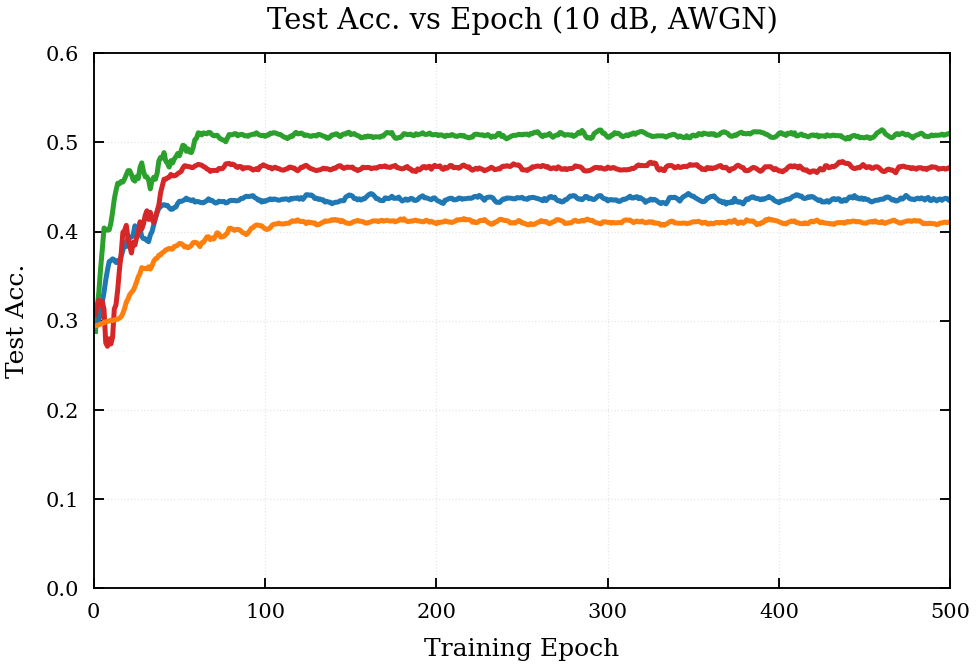}
    }\hspace{0.04\textwidth}
    \subfloat[(d) Few labels, $\text{SNR}= 20$ dB]{
        \includegraphics[width=0.31\textwidth]{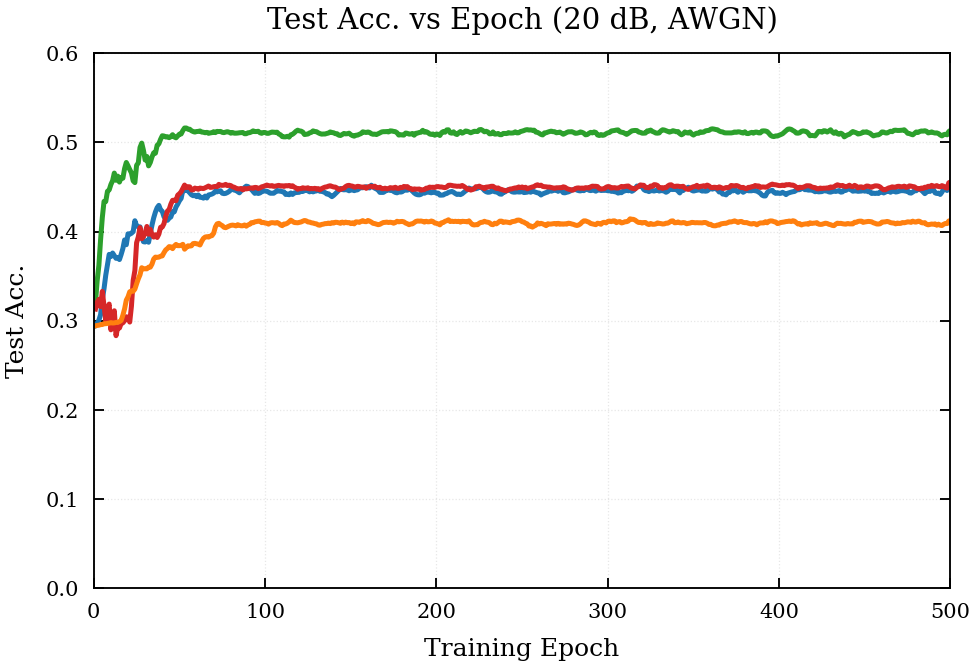}
    }

    \par\vspace{0.5ex}   

    \includegraphics[width=0.55\textwidth]{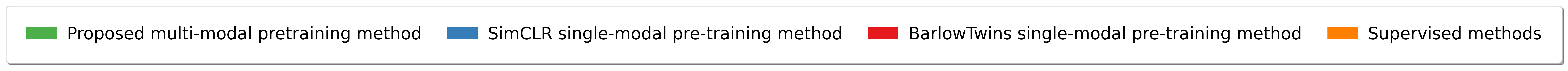}

    \vspace{-2mm}
    \caption{NYUDv2 test accuracy versus communication rounds with full and few labels.}
    \label{fig:results1}
\end{figure*}

\begin{figure*}[t]
    \centering
    \captionsetup[subfigure]{labelformat=empty,font=footnotesize}
    
    \subfloat[(a) Full labels, $\text{SNR}= 10$ dB]{
        \includegraphics[width=0.31\textwidth]{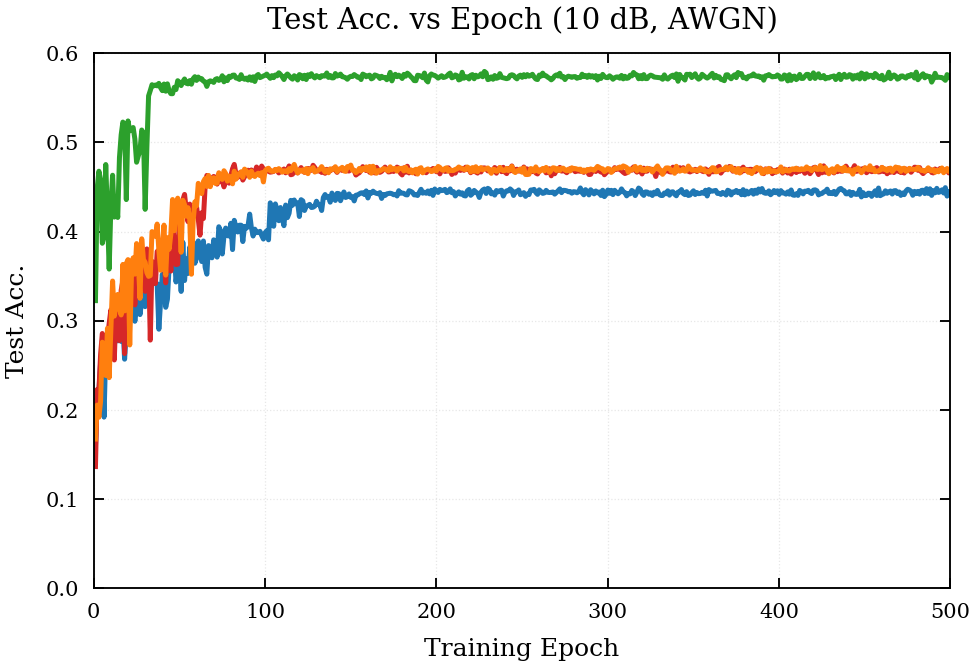}
    }\hspace{0.04\textwidth}
    \subfloat[(b) Full labels, $\text{SNR}= 20$ dB]{
        \includegraphics[width=0.31\textwidth]{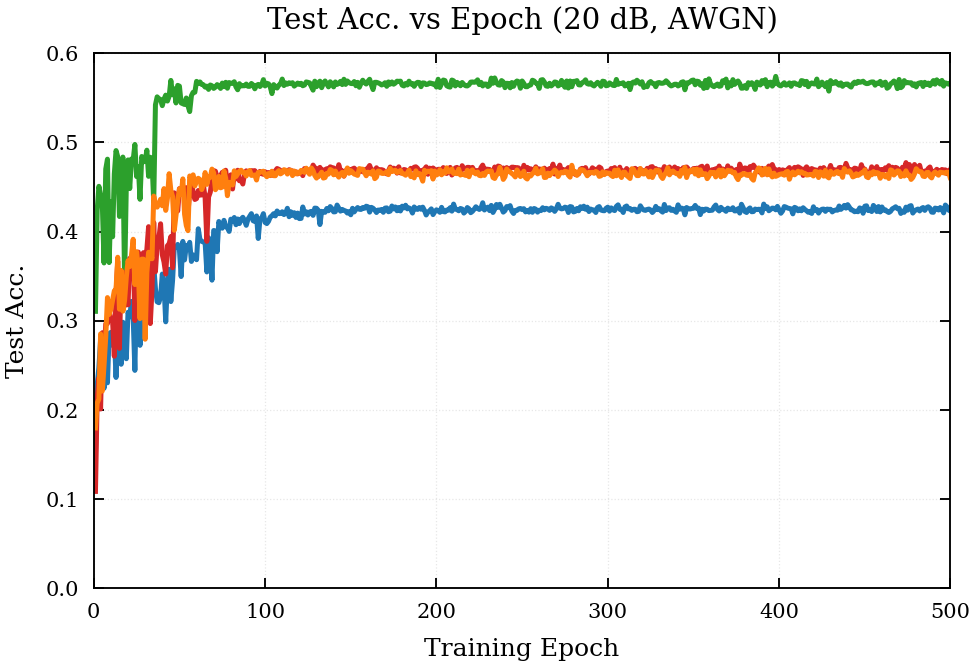}
    }

    \par\vspace{0.5ex}  

    \subfloat[(c) Few labels, $\text{SNR}= 10$ dB]{
        \includegraphics[width=0.31\textwidth]{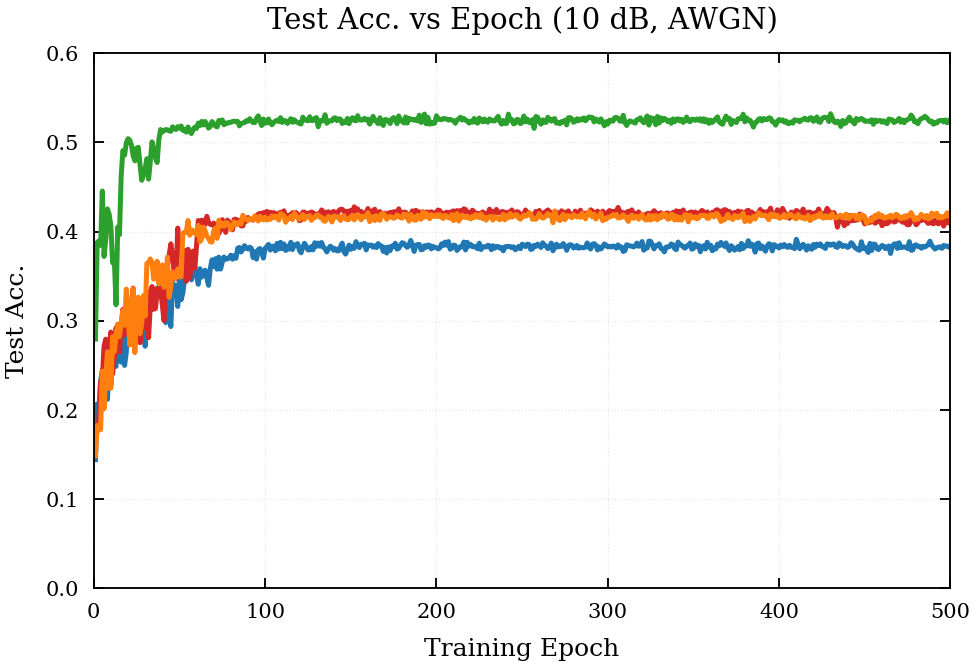}
    }\hspace{0.04\textwidth}
    \subfloat[(d) Few labels, $\text{SNR}= 20$ dB]{
        \includegraphics[width=0.31\textwidth]{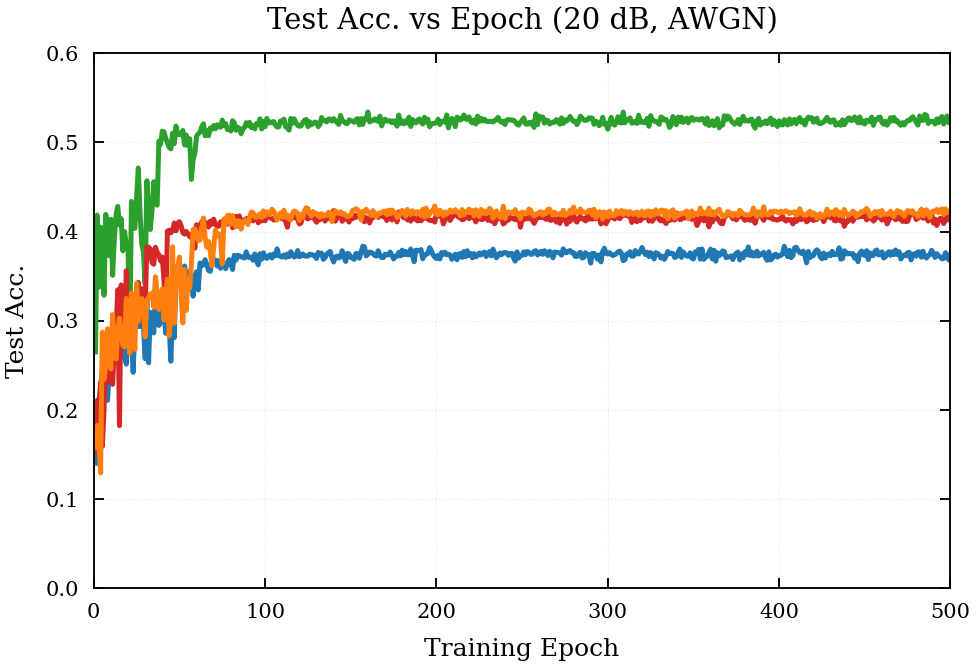}
    }

    \par\vspace{0.5ex}  

    \includegraphics[width=0.55\textwidth]{tag.png}

    \vspace{-2mm}
    \caption{SUN RGB-D test accuracy versus communication rounds with full and few labels.}
    \label{fig:results2}
\end{figure*}

Figures~\ref{fig:results1} and \ref{fig:results2} show the test accuracy on NYUDv2 and SUN RGB-D versus training epochs (i.e., communication rounds) under AWGN channels with $\mathrm{SNR}{=}10$ dB and $20$ dB. In our synchronous edge--server setup, each epoch corresponds to one full communication round, so the horizontal axis directly reflects the training-phase communication cost $C_{\text{train}}$.

For both datasets and SNRs, the full-label results in subplots (a)–(b) indicate that the proposed multi-modal pre-training converges faster and reaches a higher accuracy plateau than SimCLR, Barlow Twins, and training from scratch. With only $50\%$ labels (subplots (c)–(d)), our method still approaches its final accuracy within a small number of epochs and shows stable training, whereas the baselines converge more slowly and fluctuate more, especially Barlow Twins. This confirms that Stage~I provides a strong, label-efficient initialization for supervised fine-tuning.

Table~\ref{tab:pretrain-results} further quantifies the communication savings by reporting the number of epochs needed to reach target accuracy levels. Since each epoch is one communication round, these numbers give an empirical measure of $C_{\text{train}}$. The proposed method attains $40$–$50\%$ accuracy within only a few epochs on both datasets, while SimCLR and Barlow Twins require an order of magnitude more epochs, and the model without pre-training often fails to reach the same targets within the training budget. Thus, Stage~I substantially reduces the training-phase communication cost for a given accuracy target and also yields the highest final accuracy.

\begin{table*}[t]
\centering
\caption{Epochs required to reach target accuracy levels under different pre-training methods and datasets. 
         ``---'' indicates the target accuracy was not reached within the training budget.}
\label{tab:pretrain-results}
\begin{tabular}{@{} ll 
    S[table-format=3.0] 
    S[table-format=3.0] 
    S[table-format=2.0,table-space-text-post={---}] 
    S[table-format=1.5] @{}}
\toprule
\textbf{Pretrain Method} & \textbf{Dataset} &
\textbf{Epochs to 40\%} & \textbf{Epochs to 45\%} &
\textbf{Epochs to 50\%} & {\textbf{Best Acc}} \\
\midrule
Proposed      & NYUDv2     &   4 &   9 &  14 & 0.5637 \\
SimCLR        & NYUDv2     &  20 & 100 & --- & 0.4531 \\
Barlow Twins   & NYUDv2     &  16 &  30 &  50 & 0.5305 \\
No\_pretrain  & NYUDv2     &  61 & 377 & --- & 0.4510 \\
\addlinespace[1.5ex]
Proposed      & SUNRGBD  &   2 &   3 &  16 & 0.5778 \\
SimCLR        & SUNRGBD  &  73 & --- & --- & 0.4492 \\
Barlow Twins   & SUNRGBD  &  46 &  65 & --- & 0.4753 \\
No\_pretrain  & SUNRGBD  &  37 &  61 & --- & 0.4757 \\
\bottomrule
\end{tabular}
\end{table*}

\vspace{0.2cm}

\subsection{Inference Robustness under Varying Channels and Noisy Inputs}

To enhance inference reliability, the proposed framework integrates evidential fine-tuning with reliability-aware fusion to provide calibrated uncertainty for each modality, while utilizing these uncertainty scores to drive an adaptive retransmission policy. We now demonstrate how this uncertainty-aware inference mechanism jointly improves robustness against varying wireless channels and noisy inputs on the SUN RGB-D dataset, all while maintaining a low inference-phase communication cost $C_{\text{infer}}$.

\subsubsection{Varying Channel Environments}
We first study robustness to varying wireless channels. All models are trained with a fixed $\mathrm{SNR}{=}10$~dB and evaluated under both (i) an AWGN channel and (ii) a Rayleigh fading channel, using the same static SNRs $\{0,10,20\}$~dB and a dynamic setting where the SNR is uniformly drawn from $[0,20]$~dB at test time.
In Table~\ref{tab:snr-results}, \emph{Baseline} is a JSCC-based scheme with Stage~I pre-training, \emph{Ours (Pre-RT)} uses Stage~I and Stage~II but no retransmissions, and \emph{Ours (Post-RT)} further enables Stage~III with an average retransmission ratio of about $10\%$.


\begin{table*}[t]
\centering
\caption{Accuracy comparison under different SNR conditions over AWGN and Rayleigh channels (same SNR settings).}
\label{tab:snr-results}
\setlength{\tabcolsep}{5pt}
\renewcommand{\arraystretch}{1.05}
\begin{tabular}{@{} l l c *{3}{S[table-format=1.4]} @{}}
\toprule
\textbf{Channel} & \textbf{SNR Mode} & \textbf{SNR (dB)} &
{\textbf{Baseline}} & {\textbf{Ours (Pre-RT)}} & {\textbf{Ours (Post-RT)}} \\
\midrule
AWGN    & dynamic & 0--20 & 0.5566 & 0.5774 & 0.5786 \\
AWGN    & static  & 0     & 0.5443 & 0.5703 & 0.5735 \\
AWGN    & static  & 10    & 0.5702 & 0.5778 & 0.5790 \\
AWGN    & static  & 20    & 0.5749 & 0.5795 & 0.5807 \\
\midrule
Rayleigh & dynamic & 0--20 & \multicolumn{1}{c}{0.4948} & 0.4980 & 0.5008 \\
Rayleigh & static  & 0     & \multicolumn{1}{c}{0.4886} & 0.4952 & 0.5027 \\
Rayleigh & static  & 10    & \multicolumn{1}{c}{0.4959} & 0.5014 & 0.5027 \\
Rayleigh & static  & 20    & \multicolumn{1}{c}{0.4965} & 0.5018 & 0.5031 \\
\bottomrule
\end{tabular}
\end{table*}

Across all channel settings, \emph{Ours (Pre-RT)} improves accuracy over \emph{Baseline}, showing that evidential heads and reliability-aware fusion already provide strong robustness and better-calibrated predictions without retransmissions. Activating Stage~III (\emph{Ours (Post-RT)}) yields a further consistent gain, especially at $0$ dB and under dynamic channels, while keeping the average retransmission ratio small. The uncertainty-guided policy selectively requests extra channel uses only for difficult samples, improving reliability under poor channels and avoiding unnecessary retransmissions when the channel is good. Thus, Stage~II and Stage~III together improve robustness to channel variations at almost the same $C_{\text{infer}}$ as one-shot transmission.

\subsubsection{Noisy Input Data}
We next examine robustness to sensing noise by adding Gaussian noise with different power levels to either the depth or RGB view. Table~\ref{tab:noise-results} compares the classification accuracy of the \emph{Baseline} and the proposed method.

\begin{table}[t]
\centering
\caption{Model performance under different noise conditions.}
\label{tab:noise-results}
\begin{tabular}{@{} l l S[table-format=3.0] S[table-format=1.4] @{}}
\toprule
\textbf{Model} & \textbf{Noisy View} & {\textbf{Noise Power}} & {\textbf{Acc}} \\
\midrule
baseline & depth & 1 & 0.1687 \\
baseline & depth & 10 & 0.1638 \\
baseline & depth & 100 & 0.1719 \\
baseline & rgb & 1 & 0.1281 \\
baseline & rgb & 10 & 0.0901 \\
baseline & rgb & 100 & 0.0882 \\
\midrule[\heavyrulewidth]
proposed & depth &   1 & 0.4447 \\
proposed & depth &  10 & 0.4583 \\
proposed & depth & 100 & 0.4589 \\
proposed & rgb   &   1 & 0.2432 \\
proposed & rgb   &  10 & 0.2722 \\
proposed & rgb   & 100 & 0.2883 \\
\bottomrule
\end{tabular}
\end{table}

When the RGB view is corrupted, \emph{Baseline} accuracy quickly drops below $0.13$ and remains around $0.09$ for stronger noise, whereas our method reaches $0.24$–$0.29$. For depth, \emph{Baseline} stays near $0.17$, while our method maintains about $0.45$ accuracy across all noise powers. This demonstrates that the proposed framework is much more resilient to sensing noise on either modality.

This robustness mainly comes from Stage~II and Stage~III. Stage~II uses evidential uncertainty and reliability-aware fusion to detect noisy views and downweight them in the final decision. Stage~III then allocates extra channel uses only to samples and modalities with high uncertainty, providing redundancy where needed without overusing bandwidth. Together with the training-efficiency gains of Stage~I, these results show that training-phase communication cost, representation calibration, and uncertainty-aware resource allocation should be jointly considered when designing next-generation edge-assisted semantic communication systems.

\section{Conclusions}
This paper investigated distributed multi-modal edge inference over bandwidth-limited wireless networks, aiming to reduce device--server communication during both training and inference while improving robustness under varying channels and noisy or degraded inputs. For that purpose, we proposed a three-stage communication-aware framework that combines fully local multi-modal self-supervised pre-training at devices, evidential supervised fusion at the edge server, and an uncertainty-guided retransmission policy during inference, jointly shaping the representations, fusion, and communication decisions.
Experiments on RGB--depth indoor scene classification under wireless channels show that this framework achieves higher accuracy than conventional separate source--channel coding and alternative multi-modal baselines, while requiring substantially fewer device--server training rounds and labeled samples. At inference, evidential fusion and uncertainty-based retransmission maintain strong robustness when channel conditions vary, the SNR is low, or one modality is severely corrupted or missing, by selectively triggering additional transmissions only for uncertain samples.
Future work includes extending the framework to sequence-based tasks and richer multi-device and multi-task settings, as well as jointly learning retransmission and scheduling policies for non-stationary environments.

\appendices

\section{Proof of Theorem~\ref{thm:mv-extension} (Information-theoretic Guarantee for Multi-modal SSL)}
\label{app:proof-mv}
\paragraph*{A. Supervised learning with a perfect channel}
When $\varepsilon_{\mathrm{c}}(\mathcal{S})=0$, the channel is information preserving.
The Markov chain $Y \rightarrow X_{\mathcal{S}} \rightarrow Z_{\mathcal{S}} \rightarrow \widehat{Z}_{\mathcal{S}}$
holds.
By the data processing inequality,
\begin{equation}
I(\widehat{Z}_{\mathcal{S}};Y)\le I(X_{\mathcal{S}};Y).
\label{eq:dpi-sup}
\end{equation}
The upper bound is achievable.
One can choose a representation that is sufficient for predicting $Y$ from $X_{\mathcal{S}}$.
Let $\widehat{Z}_{\mathcal{S},\mathrm{sup}}^{\mathrm{opt}}$ be any maximizer of $I(\widehat{Z}_{\mathcal{S}};Y)$.
Let $\widehat{Z}_{\mathcal{S},\mathrm{sup}}^{\mathrm{opt,min}}$ be the minimizer of $H(\widehat{Z}_{\mathcal{S}}|Y)$ among these maximizers.
Then
\[
I(\widehat{Z}_{\mathcal{S},\mathrm{sup}}^{\mathrm{opt,min}};Y)
= \max_{\widehat{Z}_{\mathcal{S}}} I(\widehat{Z}_{\mathcal{S}};Y)
= I(X_{\mathcal{S}};Y),
\]
which proves the supervised statement in Theorem~\ref{thm:mv-extension}.

\paragraph*{B. Self-supervised learning under channel loss}
In multi-modal SSL, the objective is to maximize $I(\widehat{Z}_{\mathcal{S}};X'_{\mathcal{S}})$.
By definition,
$\varepsilon_{\mathrm{c}}(\mathcal{S})
\triangleq I(Z_{\mathcal{S}};X'_{\mathcal{S}})-I(\widehat{Z}_{\mathcal{S}};X'_{\mathcal{S}})\ge 0.$
Hence, for any pair $(Z_{\mathcal{S}},\widehat{Z}_{\mathcal{S}})$ on the channel,
\begin{equation}
I(\widehat{Z}_{\mathcal{S}};X'_{\mathcal{S}})
= I(Z_{\mathcal{S}};X'_{\mathcal{S}})-\varepsilon_{\mathrm{c}}(\mathcal{S}).
\label{eq:channel-loss}
\end{equation}
Moreover, since $Z_{\mathcal{S}}$ is generated from $X_{\mathcal{S}}$, we have the Markov chain
$X'_{\mathcal{S}} \leftarrow Y \rightarrow X_{\mathcal{S}} \rightarrow Z_{\mathcal{S}}.$
By the data processing inequality,
\begin{equation}
I(Z_{\mathcal{S}};X'_{\mathcal{S}})\le I(X_{\mathcal{S}};X'_{\mathcal{S}}).
\label{eq:dpi-ssl}
\end{equation}
Combining~\eqref{eq:channel-loss} and~\eqref{eq:dpi-ssl}, the SSL objective satisfies
\[
I(\widehat{Z}_{\mathcal{S}};X'_{\mathcal{S}})
\le I(X_{\mathcal{S}};X'_{\mathcal{S}})-\varepsilon_{\mathrm{c}}(\mathcal{S}).
\]
Let $\widehat{Z}_{\mathcal{S},\mathrm{ssl}}^{\mathrm{opt}}$ be an optimizer of
$\max_{\widehat{Z}_{\mathcal{S}}} I(\widehat{Z}_{\mathcal{S}};X'_{\mathcal{S}})$.
Let $\widehat{Z}_{\mathcal{S},\mathrm{ssl}}^{\mathrm{opt,min}}$ be the minimizer of $H(\widehat{Z}_{\mathcal{S}}|Y)$ among the optimizers.
Then
\begin{equation}
I(\widehat{Z}_{\mathcal{S},\mathrm{ssl}}^{\mathrm{opt,min}};X'_{\mathcal{S}})
= I(\widehat{Z}_{\mathcal{S},\mathrm{ssl}}^{\mathrm{opt}};X'_{\mathcal{S}})
= I(X_{\mathcal{S}};X'_{\mathcal{S}})-\varepsilon_{\mathrm{c}}(\mathcal{S}).
\label{eq:opt-ssl}
\end{equation}

\paragraph*{C. Lower bound on task-relevant information}
From the graphical model~\eqref{eq:mv-pgm}, conditioning on $Y$ blocks all paths between
$\widehat{Z}_{\mathcal{S}}$ and $X'_{\mathcal{S}}$.
Therefore,
\begin{equation}
I(\widehat{Z}_{\mathcal{S}};X'_{\mathcal{S}}\mid Y)=0.
\label{eq:cond-indep}
\end{equation}
Apply the chain rule to $I(\widehat{Z}_{\mathcal{S}};Y,X'_{\mathcal{S}})$ in two ways:
\begin{align}
I(\widehat{Z}_{\mathcal{S}};Y,X'_{\mathcal{S}})
&= I(\widehat{Z}_{\mathcal{S}};X'_{\mathcal{S}})+I(\widehat{Z}_{\mathcal{S}};Y\mid X'_{\mathcal{S}}),
\label{eq:chain-1}\\
I(\widehat{Z}_{\mathcal{S}};Y,X'_{\mathcal{S}})
&= I(\widehat{Z}_{\mathcal{S}};Y)+I(\widehat{Z}_{\mathcal{S}};X'_{\mathcal{S}}\mid Y).
\label{eq:chain-2}
\end{align}
Using~\eqref{eq:cond-indep} in~\eqref{eq:chain-2} and comparing with~\eqref{eq:chain-1}, we obtain
\begin{equation}
I(\widehat{Z}_{\mathcal{S}};Y)
= I(\widehat{Z}_{\mathcal{S}};X'_{\mathcal{S}})
+ I(\widehat{Z}_{\mathcal{S}};Y\mid X'_{\mathcal{S}})
\ge I(\widehat{Z}_{\mathcal{S}};X'_{\mathcal{S}}).
\label{eq:yz-lb}
\end{equation}
Apply~\eqref{eq:yz-lb} to $\widehat{Z}_{\mathcal{S},\mathrm{ssl}}^{\mathrm{opt,min}}$ and use~\eqref{eq:opt-ssl}:
\begin{equation}
I(\widehat{Z}_{\mathcal{S},\mathrm{ssl}}^{\mathrm{opt,min}};Y)
\ge I(X_{\mathcal{S}};X'_{\mathcal{S}})-\varepsilon_{\mathrm{c}}(\mathcal{S}).
\label{eq:lb-step1}
\end{equation}

It remains to express $I(X_{\mathcal{S}};X'_{\mathcal{S}})$ in terms of $I(X_{\mathcal{S}};Y)$.
From~\eqref{eq:mv-pgm}, we also have $I(X_{\mathcal{S}};X'_{\mathcal{S}}\mid Y)=0$.
Use the chain rule for $I(X_{\mathcal{S}};Y,X'_{\mathcal{S}})$:
\begin{align}
I(X_{\mathcal{S}};Y,X'_{\mathcal{S}})
&= I(X_{\mathcal{S}};X'_{\mathcal{S}})+I(X_{\mathcal{S}};Y\mid X'_{\mathcal{S}}), \label{eq:chain-X-1}\\
&= I(X_{\mathcal{S}};Y)+I(X_{\mathcal{S}};X'_{\mathcal{S}}\mid Y). \label{eq:chain-X-2}
\end{align}

The last term is zero, so
\begin{equation}
I(X_{\mathcal{S}};X'_{\mathcal{S}})
= I(X_{\mathcal{S}};Y)-I(X_{\mathcal{S}};Y\mid X'_{\mathcal{S}}).
\label{eq:xy-identity}
\end{equation}
Substitute~\eqref{eq:xy-identity} into~\eqref{eq:lb-step1}:
\[
I(\widehat{Z}_{\mathcal{S},\mathrm{ssl}}^{\mathrm{opt,min}};Y)
\ge I(X_{\mathcal{S}};Y)
- I(X_{\mathcal{S}};Y\mid X'_{\mathcal{S}})
- \varepsilon_{\mathrm{c}}(\mathcal{S}).
\]
This is the desired lower bound.

\paragraph*{D. Upper bound and the sandwich inequality}
From $Y \rightarrow X_{\mathcal{S}} \rightarrow \widehat{Z}_{\mathcal{S}}$, the data processing inequality gives
\[
I(\widehat{Z}_{\mathcal{S},\mathrm{ssl}}^{\mathrm{opt,min}};Y)\le I(X_{\mathcal{S}};Y).
\]
Together with the lower bound above, we obtain~\eqref{eq:mv-bound-main}.
The proof is complete.


\bibliographystyle{IEEEtran}
\bibliography{reference}

\end{document}